# Local search for stable marriage problems with ties and incomplete lists

M. Gelain[1], M. S. Pini[1], F. Rossi[1], K. B. Venable[1], T. Walsh[2]

[1] Dipartimento di Matematica Pura ed Applicata, Università di Padova, Italy
E-mail: {mgelain,mpini,frossi,kvenable}@math.unipd.it
[2] NICTA and UNSW Sydney, Australia,
Email: Toby.Walsh@nicta.com.au

**Abstract.** The stable marriage problem has a wide variety of practical applications, ranging from matching resident doctors to hospitals, to matching students to schools, or more generally to any two-sided market. We consider a useful variation of the stable marriage problem, where the men and women express their preferences using a preference list with ties over a subset of the members of the other sex. Matchings are permitted only with people who appear in these preference lists. In this setting, we study the problem of finding a stable matching that marries as many people as possible. Stability is an envy-free notion: no man and woman who are not married to each other would both prefer each other to their partners or to being single. This problem is NP-hard. We tackle this problem using local search, exploiting properties of the problem to reduce the size of the neighborhood and to make local moves efficiently. Experimental results show that this approach is able to solve large problems, quickly returning stable matchings of large and often optimal size.

## 1 Introduction

The stable marriage problem [1] is a well-known problem of matching men to women to achieve a certain type of "stability". Each person expresses a strict preference ordering over the members of the opposite sex. The goal is to match men to women so that there are no two people of opposite sex who would both rather be matched with each other than with their current partners. Surprisingly such a stable marriage always exists and one can be found in polynomial time. Gale and Shapley give a quadratic time algorithm to solve this problem based on a series of proposals of the men to the women (or vice versa) [2]. The stable marriage problem has a wide variety of practical applications, ranging from matching resident doctors to hospitals, sailors to ships, primary school students to secondary schools, as well as in market trading.

There are many variants of this classical formulation of the stable marriage problem. Some of the most useful in practice include incomplete preference lists (SMI), that allows us to model unacceptability for certain members of the other sex, and preference lists with ties (SMT), that model indifference in the preference ordering. With a SMI problem, we have to find a stable marriage in which the married people accept each other. It is known that all solutions of a SMI problem have the same size [3] (that is, number of married people). In SMT problems, instead, solutions are stable marriages where everybody is married. Both of these variants are polynomial to solve. In real world situations, both ties and incomplete preference lists may be needed. Unfortunately, when we allow both, the problem becomes NP-hard [3]. In a SMTI (Stable Marriage with Ties and Incomplete lists) problem, there may be several stable marriages of different sizes, and solving the problem means finding a stable marriage of maximum size.

In this paper we investigate the use of a local search approach to tackle this problem. Our algorithm starts from a randomly chosen marriage and, at each step, moves to a neighbor marriage which is obtained by removing one blocking pair, that is, a man-woman pair who are not married to each other in the current marriage but who prefer to be married with each other rather than with with their current partners. Stable marriages have no blocking pairs, so the aim of such a move is to pass to a marriage which is closer to stability. Among the neighbor marriages, the evaluation function chooses one with the smallest number of blocking pairs and of singles. Since there may be several stable marriages with different sizes, we look for the one with maximum size (that is, the smallest number of singles). Random moves are also used, to avoid stagnation in local minima. The algorithm stops when a perfect matching (that is, a stable marriage with no singles) is found, or when a given limit on the number of steps is reached.

This basic local search approach works well with problems of limited size, but does not scale. With large sizes, it fails to find good solutions and sometimes even stable marriages. One of the main reasons is that the neighborhood can be very large, since a marriage may have a large number of blocking pairs. Many such blocking pairs can be ignored since they are "dominated" by others, whose removal will also eliminate all the dominated blocking pairs. By considering only undominated blocking pairs, we can solve SMTI problems of much larger size in a small amount of time. The marriages returned by our local search method are stable and contain very few single people. Experiments on randomly generated SMTI problems of size 100 show that our algorithm is able to find stable marriages with at most two singles on average in tens of seconds at worst.

The SMTI problem has been tackled also in [4], where the problem is modeled as a constraint optimization problem and a constraint solver is employed to solve it. This systematic approach is guaranteed to find always an optimal solution. However, our experimental results show that our local search algorithm in practice always finds optimal solutions. Moreover, it scales well to sizes much larger than those considered in [4]. Instances of size comparable to ours are considered in [5]. However, the problem solved in that paper is the decision version of our optimization problem. That is, they ask if there exists a stable marriage of a certain size. Another approach is to use approximation. Given an SMTI problem, if its maximum cardinality stable marriage marriages are of size $k$, an $\alpha/\beta$-approximation algorithm is able to return a stable marriage of size at least $\beta/\alpha \cdot k$. The SMTI problem cannot have and $i$-approximation algorithm for $i$ greater than 33/29 unless P=NP [6]. A 3/2-approximation algorithm has been proposed in [7].

## 2 Background

### 2.1 Stable marriage problems with ties and incompleteness

A stable marriage (SM) problem [1] consists of matching members of two different sets, usually called men and women. When there are $n$ men and $n$ women, the SM problem is said to have size $n$. Each person strictly ranks all members of the opposite sex. The goal is to match the men with the women so that there are no two people of opposite sex who would both rather marry each other than their current partners. Such a marriage is called *stable.* At least one stable marriage exists for every SM problem. In fact, the set of stable marriages forms a lattice. Gale and Shapley give a polynomial time algorithm to find the stable marriage at the top (or bottom) of this lattice [2].

In this paper we consider a variant of the SM problem where preference lists may include ties and may be incomplete. This variant is denoted by SMTI [8]. Ties express indifference in the preference ordering, while incompleteness models unacceptability for certain partners.

**Definition 1 (SMTI marriage).** *Given a SMTI problem with $n$ men and $n$ women, a marriage $M$ is a one-to-one matching between men and women such that partners are acceptable for each other. If a man $m$ and a woman $w$ are matched in $M$, we write $M(m) = w$ and $M(w) = m$. If a person $p$ is not matched in $M$ we say that he/she is single.*

**Definition 2 (Marriage size).** *Given a SMTI problem of size $n$ and a marriage $M$, its size is the number of men (or women) that are married.*

An example of a SMTI problem with four men and women is shown in Table 1. A SMTI problem is described by giving, for each man and woman, the corresponding preference list over members of the other sex. For example, by writing 2 : 2 (3 4) among the men's preference lists we mean that man $m_2$ strictly prefers woman $w_2$ to women $w_3$ and $w_4$, that are equally preferred.

| men's preference lists | women's preference lists |
|---|---|
| 1: 2 1 | 1: 3 1 (2 4) |
| 2: 2 (3 4) | 2: 1 4 2 |
| 3: (1 2 3 4) | 3: (1 2) (4 3) |
| 4: (3 2) 1 4 | 4: (3 2 4) |

**Table 1.** An example of a SMTI problem of size 4.

**Definition 3 (Blocking pairs in SMTIs).** *Consider a SMTI problem $P$, a marriage $M$ for $P$, a man $m$ and a woman $w$. A pair $(m, w)$ is a blocking pair* *in $M$ if $m$ and $w$ find acceptable each other and $m$ is either single in $M$ or he strictly prefers $w$ to $M(m)$, and $w$ is either single in $M$ or she strictly prefers $m$ to $M(w)$.*

**Definition 4 (Weakly Stable Marriages).** *Given a SMTI problem $P$, a marriage $M$ for $P$ is weakly stable if it has no blocking pairs.*

As we will consider only weakly stable marriages, we will simply call them stable marriages. Given a SMTI problem, there may be several stable marriages of different size. If the size of a marriage coincides with the size of the problem, it is said to be a perfect matching.

In the above example, the marriage 2 3 1 4 (where the number in position $i$ indicates the woman married to man $m_i$ in that marriage) is stable and its size is 4, so it is a perfect matching.

Solving a SMTI problem means finding a stable marriage with maximal size. This problem is NP-hard [3].

### 2.2 Local search

Local search [9, 10] is one of the fundamental paradigms for solving computationally hard combinatorial problems. Local search methods in many cases represent the only feasible way for solving large and complex instances. Moreover, they can naturally be used to solve optimization problems.

Given a problem instance, the basic idea underlying local search is to start from an initial search position in the space of all solutions (typically a randomly or heuristically generated candidate solution, which may be infeasible, sub-optimal or incomplete), and to improve iteratively this candidate solution by means of typically minor modifications. At each *search step* we move to a position selected from a *local neighborhood*, chosen via a heuristic evaluation function. The evaluation function typically maps the current candidate solution to a real number and it is such that its global minima correspond to solutions of the given problem instance. The algorithm moves to the neighbor with the smallest value of the evaluation function.

This process is iterated until a *termination criterion* is satisfied. The termination criterion is usually the fact that a solution is found or that a predetermined number of steps is reached, although other variants may stop the search after a predefined amount of time.

Different local search methods vary in the definition of the neighborhood and of the evaluation function, as well as in the way in which situations are handled when no improvement is possible. To ensure that the search process does not stagnate in unsatisfactory candidate solutions, most local search methods use randomization: at every step, with a certain probability a random move is performed rather than the usual move to the best neighbor.

## 3 Local search on SMTIs

We adapt the classical local search schema to SMTI problems as follows. Given a SMTI problem $P$, we start from a randomly generated marriage $M$ for $P$. At each search step, we move to a new marriage in the neighborhood of the current one. For each marriage $M$, the neighborhood $N(M)$ is the set of all marriages obtained by removing one blocking pair from $M$. Consider a blocking pair $bp = (m, w)$ in $M$ and assume $m' = M(w)$ and $w' = M(m)$. Then, removing $bp$ from $M$ means obtaining a marriage $M'$ in which $m$ is married with $w$ and both $m'$ and $w'$ become single, leaving the other pairs in the marriage $M$ unchanged. Notice that, if $M$ is stable, its neighborhood is empty. Notice also that this notion of neighborhood is not symmetric.

To select the neighbor to move to, we use an evaluation function $f : \mathcal{M}_n \rightarrow Z$, where $\mathcal{M}_n$ is the set of all possible marriages of size $n$, and $f(M) = nbp(M) + ns(M)$. For each marriage $M$, $nbp(M)$ is the number of blocking pairs in $M$, while $ns(M)$ is the number of singles in $M$ which are not in any blocking pair. The algorithm moves to a marriage $M' \in N(M)$ such that $f(M') \leq f(M'')$ $\forall M'' \in N(M)$.

During the search, the algorithm maintains the best marriage found so far, defined as follows: if no stable marriage has been found, then the best marriage is the one with the smallest value of the evaluation function; otherwise, it is the stable marriage with less singles.

To avoid stagnation in a local minimum of the evaluation function, at each search step we perform a random walk with probability $p$ (where $p$ is a parameter of the algorithm). In the random walk, we move to a randomly selected marriage in the neighborhood (we tried also to move to a generic random marriage, but this gave worse behavior). If a stable marriage is reached, its neighborhood is empty and a random restart is performed.

The algorithm terminates if a perfect marriage (that is, a stable marriage with no singles) is found, or when a maximal number of search steps is reached. Upon termination, the algorithm returns the best marriage found during the search.

The pseudo-code of our algorithm, called LTI, is shown in Algorithm 1. In the pseudo-code, $M_{best}$ is the best marriage found so far, and $f_{best}$ its evaluation (number of blocking pairs plus number of singles). Function *best_neighbor* returns one of the best marriages in the neighborhood of the current marriage, according to the evaluation function.

In addition to this simple local search algorithm which directly applies standard local search approaches to SMTI problems, we have also designed a more sophisticated algorithm which has been tailored to exploit the specific features of SMTI problems. The main difference is in the definition of the neighborhood, which refers to the notion of *undominated* blocking pairs.

**Definition 5 (Dominance in blocking pairs).** *Let $(m, w)$ and $(m, w')$ be two blocking pairs. Then $(m, w)$ dominates (from the men's point of view) $(m, w')$ if $m$ prefers $w$ to $w'$. There is an equivalent concept from the women's point of view.*

**Definition 6 (Undominated blocking pair).** *A men- (resp., women-) undominated blocking pair is a blocking pair such that there is no other blocking* *pair that dominates it from the men's (resp., women's) point of view. When the point of view (men or women) is clear or not important, we will omit it.*

**Algorithm 1**: LTI

```
input  : a SMTI problem P, an integer max_steps, a probability p
output: a marriage
M ← random marriage
steps ← 0
M_best ← M
f_best ← f(M)
repeat
    if f(M) = 0 then
        return M
    if rand() ≤ p then
        M ← RandomWalk(M)
    else
        PAIRS ← blocking pairs in M
        if PAIRS is empty then
            perform a random restart
        else
            M ← best_neighbor(M, PAIRS)
    if M is the first stable marriage found so far then
        f_best ← f(M), M_best ← M
    if M_best is not stable and f_best > f(M) then
        f_best ← f(M), M_best ← M
    if both M_best and M are stable and f_best > f(M) then
        f_best ← f(M), M_best ← M
    steps ← steps + 1
until steps > max_steps ;
return M_best
```

For example, consider the SMTI problem in Table 1, the marriage 1 2 3 4, and two blocking pairs $(m_1, w_2)$ and $(m_4, w_2)$. Using the definitions above, $(m_1, w_2)$ dominates $(m_4, w_2)$ from the women's point of view. If we remove $(m_4, w_2)$ from the marriage, $(m_1, w_2)$ will remain. On the other hand, removing $(m_1, w_2)$ also eliminates $(m_4, w_2)$. Thus removing undominated blocking pairs may reduce the number of blocking pairs more than eliminating dominated pairs.

We call LTIU the algorithm LTI where the neighborhood is defined as the set of marriages obtained from the current one by removing any dominated blocking pair. More precisely, at each step we consider the undominated blocking pairs from the men's point of view which are also undominated from women's point of view. Notice that, in this step, the role of men and women matters, and will yield a different result if swapped.

Then, to ensure gender neutrality in our algorithm[3], in the next step we swap genders and do the same.

Due to their ability to restart, our algorithms have the PAC (probabilistically approximate complete property) [11]. That is, as their runtime goes to infinity, the probability that the algorithm returns an optimal solution goes to one. If the algorithm starts at a stable marriage, the algorithms will perform a random restart, which will end up in an optimal solution with probability greater than zero. On the other hand, if the algorithm starts from a non-stable marriage, we perform one or more steps in which we remove a blocking pair. This sequences of blocking pair removal have been shown to converge to a stable marriage with non-zero probability in the context of SMs with incomplete preference lists [12]. The proof of this result can be adapted to our context, as we have ties in the preference lists. Since a stable marriage can be reached with non-zero probability, and as we have argued above that from any stable marriage random restarting will reach an optimal solution with non-zero probability, the PAC property holds.

## 4 Experimental setting

Problems are generated using the same method as in [4]. The generator takes three parameters: the problem's size $n$, the probability of incompleteness $p_1$ and the probability of ties $p_2$. Given a triple $(n, p_1, p_2)$, a SMTI problem with $n$ men and $n$ women is generated, as follows:

1. For each man and woman, we generate a random preference list of size $n$, i.e., a permutation of $n$ persons;
2. We then iterate over each man's preference list: for a man $m_i$ and for each women $w_j$ in his preference list, with probability $p_1$ we delete $w_j$ from $m_i$'s preference list and delete $m_i$ from $w_j$'s preference list. In this way we get a possibly incomplete preference list.
3. If any man or woman has an empty preference list, we discard the problem and go to step 1.
4. We iterate over each person's (men and women's) preference list as follows: for a man $m_i$ and for each woman in his preference list, in position $j \geq 2$, with probability $p_2$ we set the preference for that woman as the preference for the woman in position $j - 1$ (thus putting the two women in a tie).

Note that this method generates SMTI problems in which the acceptance is symmetric. In fact, if a woman $w$ is not acceptable for a man $m$, $m$ is removed from $w$'s preference list. This does not introduce any loss of generality because, even if such a removal is not performed, $m$ and $w$ cannot be matched together in any stable marriage.

Notice also that this generator will not construct a SMTI problem in which a man (resp., woman) has in his preference list only women (resp., men) who do not find him (resp, her) acceptable. Such a man (resp., woman) will remain single in every stable matching. Therefore a simple preprocessing step can remove such men and women, giving a smaller problem of the form constructed by our generator.

[3] Gender neutrality is usually considered a desirable feature in a stable marriage procedure.

We generated random SMTI problems of size 100, by letting $p_2$ vary in [0, 1.0] with step 0.1, and $p_1$ vary in [0.1, 0.8] with step 0.1 (above 0.8 the preference lists start to be empty). For each parameter combination, we generated 100 problem instances. Moreover, the probability of the random walk is set to $p$=20% and the search step limit is $s$=50000.

### 4.1 Experimental results

We run our experiments on 2 x Quad-Core AMD Opteron 2.3GHz CPU with 2GB of RAM. In practice we used only one core because our algorithm is not designed for multi threading.

We first analyzed the behavior of the base algorithm, LTI. Unfortunately this algorithm fails to find a stable marriage in most of our test problems (see Figure 1). In fact, LTI always finds a stable marriage for problems where there are many ties (that is, $p_2$ high) and/or a lot of incompleteness (that is, $p_1$ high).

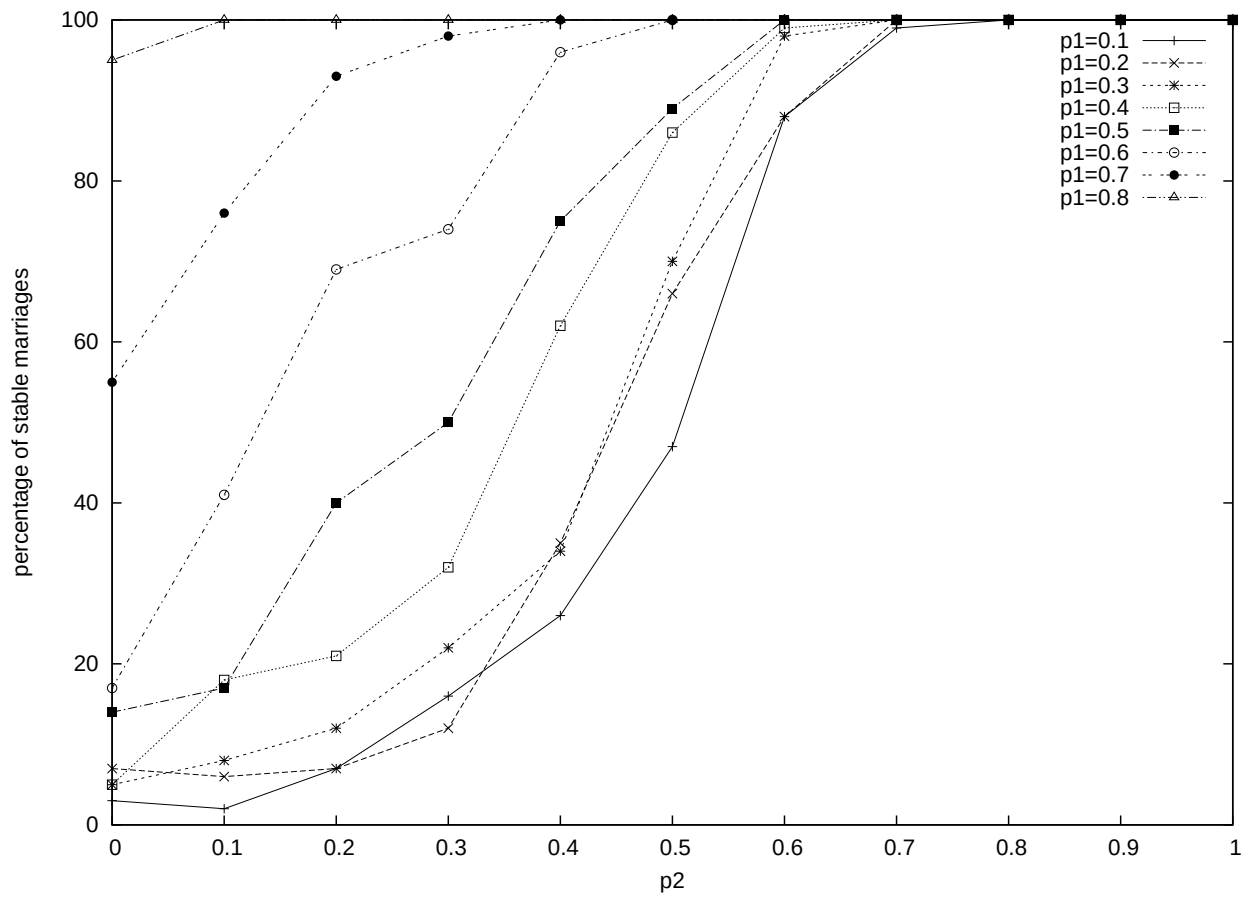

**Fig. 1.** Average number of stable marriages found by LTI.

On the other hand, algorithm LTIU finds a stable marriage in 100% of the runs. Since stability is essential in our context, from now on we will only show the experimental results for algorithm LTIU.

We start by showing the average size of the marriages returned by LTIU. In Figure 2 we can see that LTIU finds a perfect marriage (that is, a stable marriage with no singles) almost always. Even in settings with a large amount of incompleteness (that is, $p1$ = 0.7 - 0.8) the algorithm finds very large marriages, with only 2 singles on average.

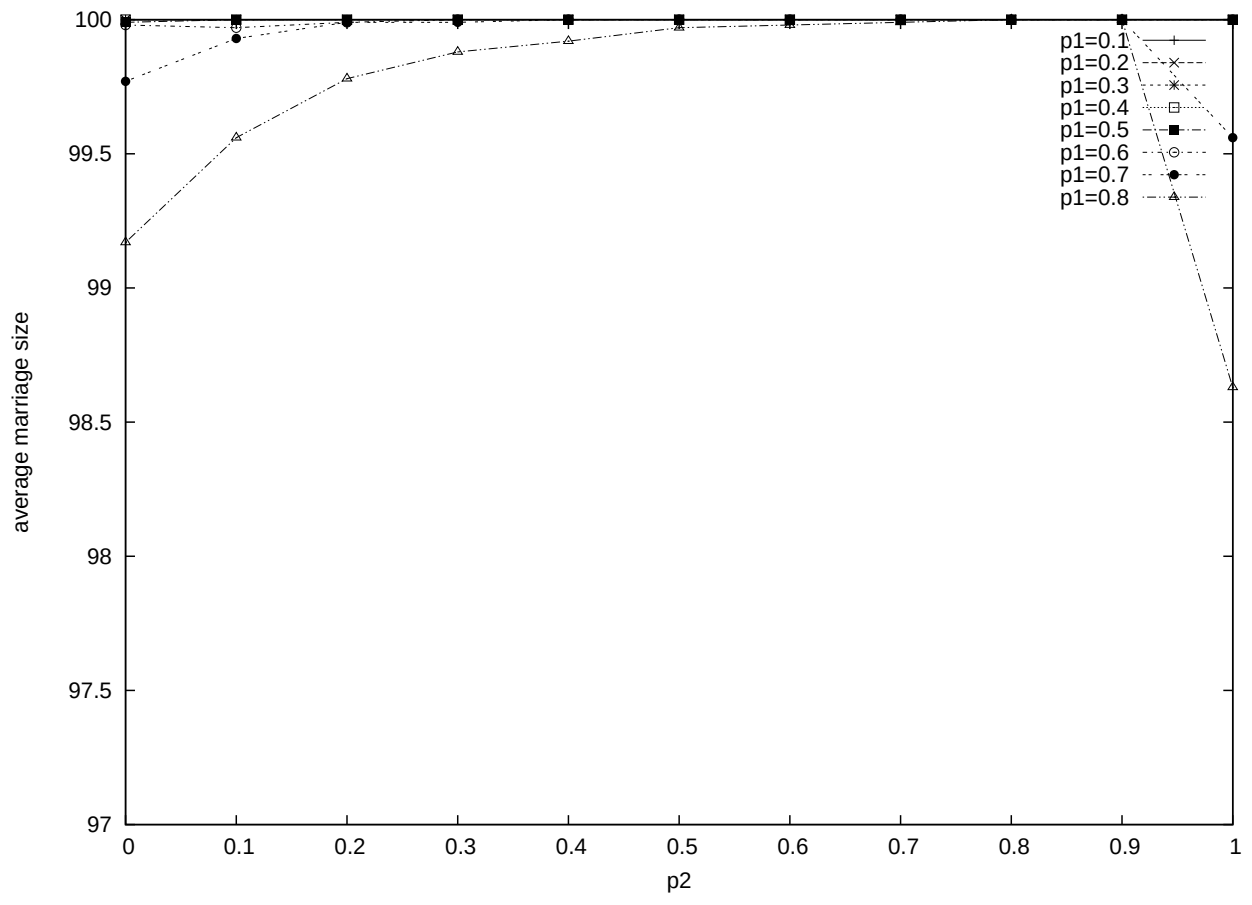

**Fig. 2.** Average size of marriages with LTIU.

We also consider the number of steps needed by our algorithm. From Figure 3(a), we can see that the number of steps is less than 2000 most of the time, except for problems with a large amount of incompleteness (i.e. $p_1 = 0.8$). As expected, with $p_1$ greater than 0.6, the algorithm requires more steps. In some cases, it reaches the step limit of 50000. Moreover, as the percentage of ties rises, stability becomes easier to achieve and thus the number of steps tends to decrease slightly. We note that complete indifference (i.e. $p_2$=1) is a special case. In fact, in this situation, the number of steps increases for almost every value of $p_1$. This is because the algorithm makes most of its progress via random restarts. In these problems every person in a preference list is equally preferred to all the others. This means that the only blocking pairs are those involving singles who both find acceptable each other. In this situation, after a few steps all singles that can be married are matched, stability is reached, and the neighborhood becomes empty. The algorithm therefore performs another random restart. It is therefore very difficult to reach a perfect matching and the algorithm often runs until the step limit.

The algorithm takes, on average, less than 40 seconds to give a result even for problems with a lot of incompleteness (see Figure 3(b)). As expected, with $p_2 = 1$ the time increases for the same reason discussed above concerning the number of steps.

Re-considering Figure 2 and the fact that all the marriages the algorithm finds are stable, we notice that most of the marriages are perfect.

From Figure 4 we see that the average percentage of matchings that are perfect is almost always 100% and this percentage only decreases when the incompleteness is large.

We compared our local search approach to the complete method from [4]. In their experiments, they measured the maximum size of the stable marriages in problems of size 10, fixing $p_1$ to 0.5 and varying $p_2$ in [0,1]. We did the same experiments (generating new instances), and obtained stable marriages of a very similar size to those reported in [4]. This means that although our algorithm is incomplete in principle, it always finds an optimal solution in our randomly generated instances, and for small sizes it behaves as a complete algorithm in terms of size of the returned marriage. However, we can also tackle problems of much larger sizes (at least 100), still obtaining optimal solutions most of the times.

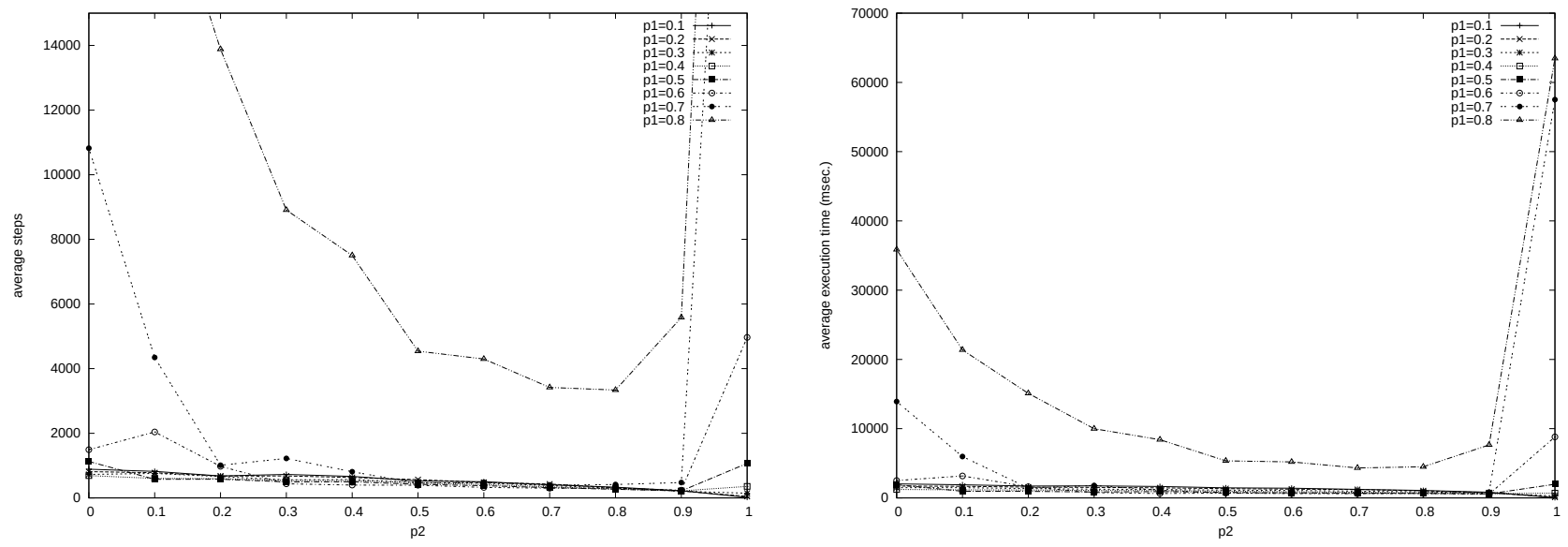

(a) average number of steps (b) average execution time

**Fig. 3.** Average number of steps and execution time for LTIU.

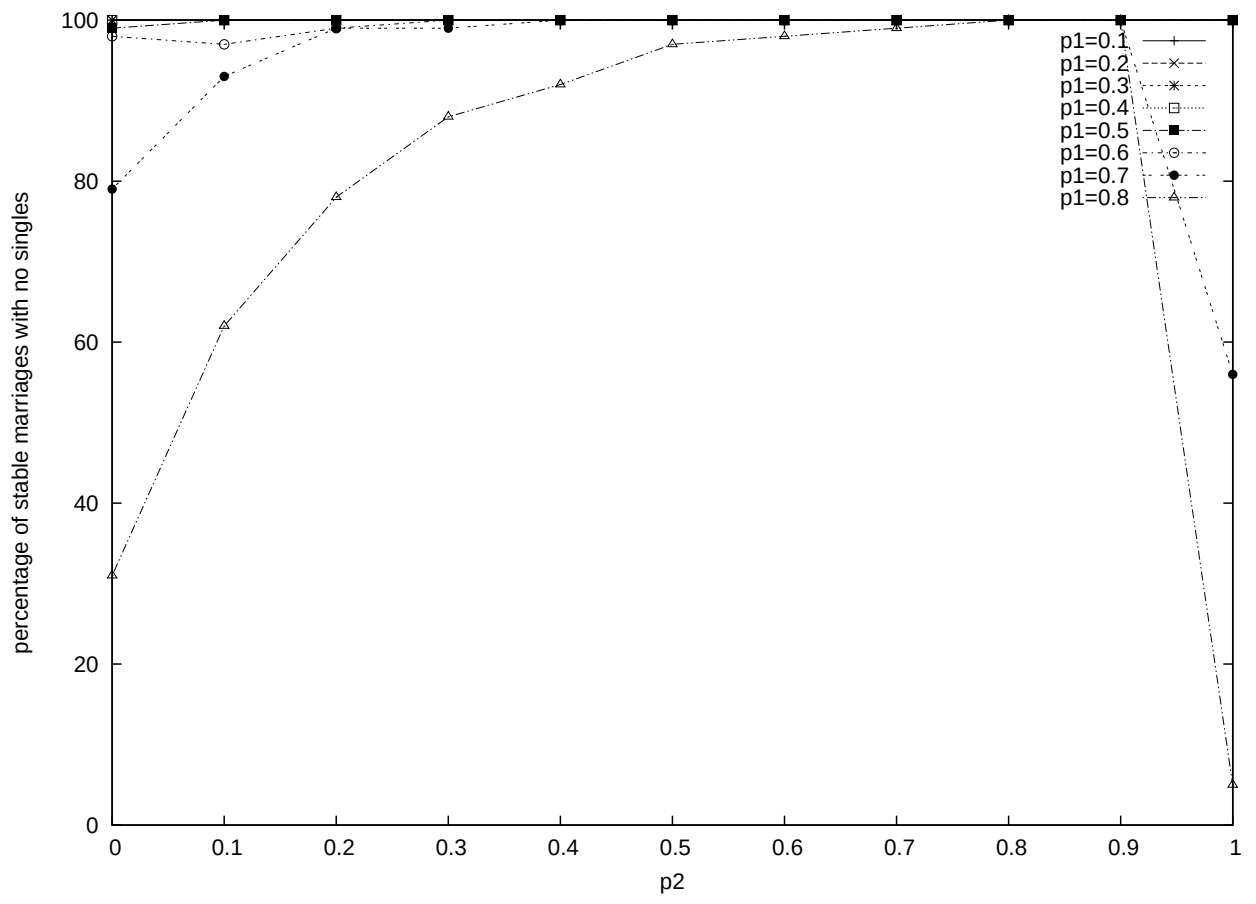

**Fig. 4.** Percentage of perfect matchings.

We also considered the runtime behavior of our algorithm. In Figure 5 we show the average normalized number of blocking pairs and, in Figure 6, the average normalized number singles of the best marriage as the execution proceeds. Although the step limit is 50000, we only plot results for the first steps because the rest is a long plateau that is not very interesting. We shows the results only for $p_2 = 0.5$. However, for greater (resp., lower) number of ties the curves are shifted slightly down (resp., up). From Figure 5 we can see that the average number of blocking pairs decreases very fast, reaching 5 blocking pairs after only 100 steps. Then, after 300-400 steps, we reach 0 blocking pairs (i.e. a stable marriage) almost all the times for all values of $p_1$. Considering Figure 6, we can see that the algorithm starts with more singles for greater values of $p_1$. This happens because, with more incompleteness, it is more unprobable for a person to be acceptable. However, after 200 steps, the average number of singles becomes very small no matter the incompleteness in the problem. Looking at both Figures 5 and 6, we observe that, although we set a step limit $s$ = 50000, the algorithm reaches a very good solution after just 300-400 steps. In fact, after this number of steps, the best marriage found by the algorithm usually has no blocking pairs nor singles, i.e. it is a perfect matching. This appears largely independent of the amount of incompleteness and the number of ties in the problems. Hence, for SMTI problems of size 100 we could set the step limit to just 400 steps and still be reasonably sure that the algorithm will return a stable marriage with a large size, no matter the amount of incompleteness and ties.

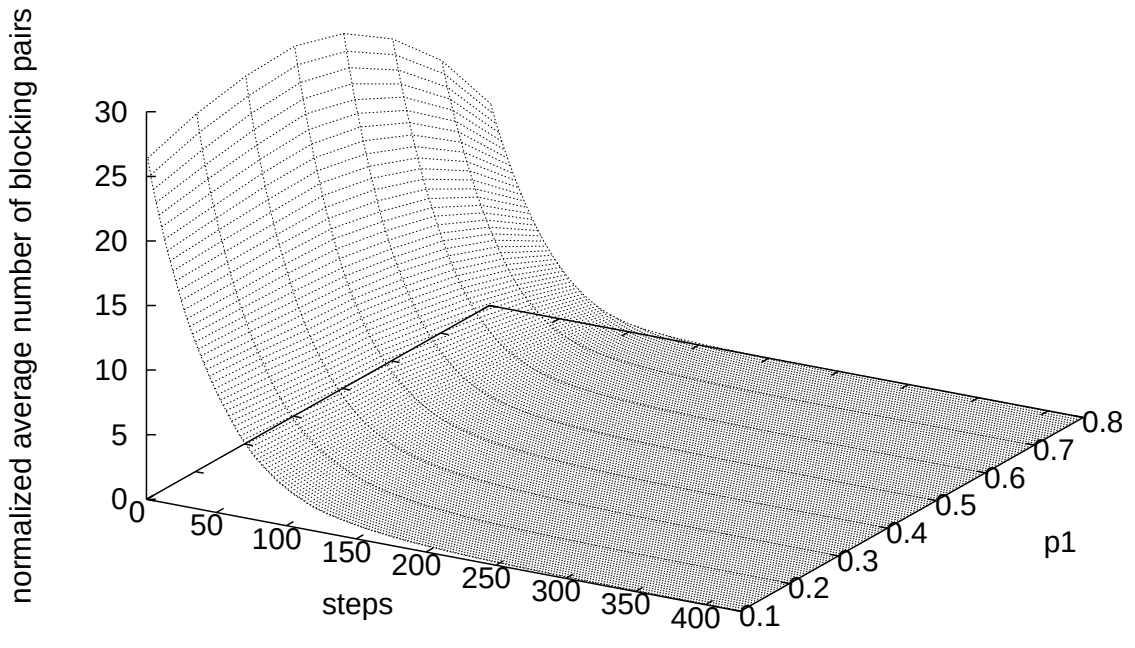

**Fig. 5.** Average normalized number of blocking pairs ($p_2$=0.5).

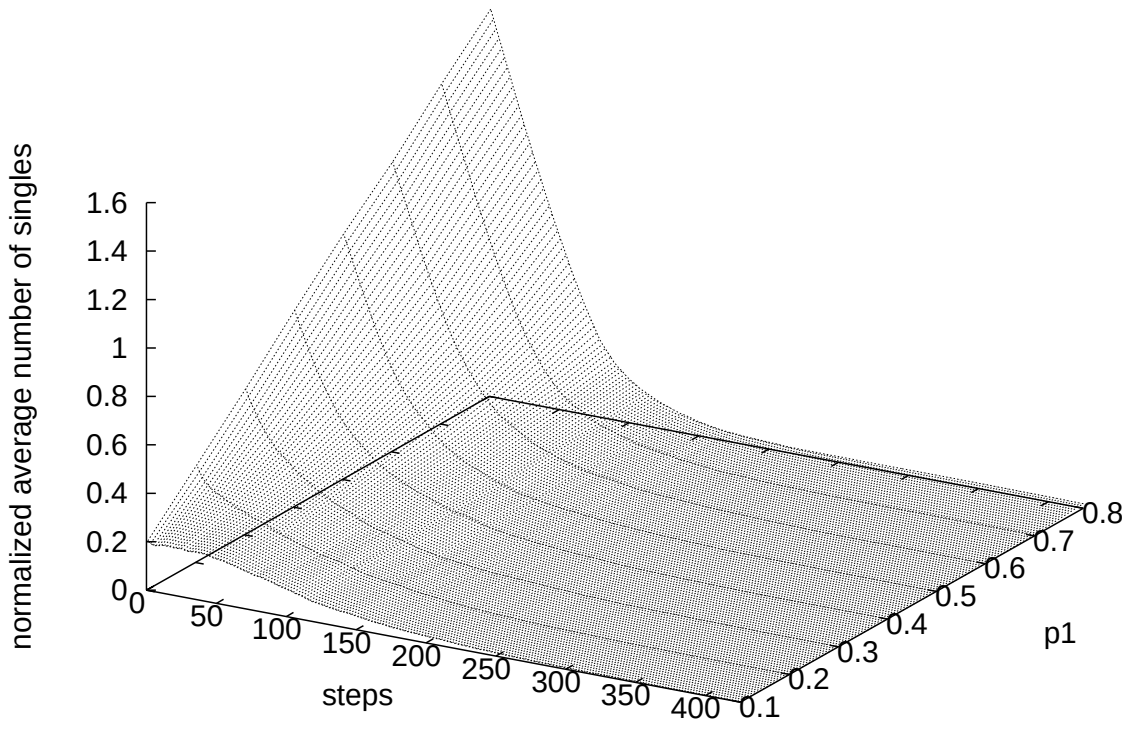

**Fig. 6.** Average normalized number of singles ($p_2$=0.5).

## 5 Conclusions

We have presented a local search approach for solving stable marriage problems with ties and indifference. Experimental results show that our algorithm is both fast and effective at finding large stable marriages .Moreover, the runtime behavior of the algorithms is not greatly influenced by the amount of incompleteness or ties in the problem. The algorithm was usually able to obtain a very good solution after a very small amount of time.

Future directions include an assessment of the trade-off between the cost of finding the undominated blocking pairs and that of treating larger neighborhoods. We also plan to apply a local search approach to other versions of the SMTI problem and to study other variant of our algorithm, for example including tabu search or other greedy heuristics.